\documentclass[conference]{IEEEtran}

\IEEEoverridecommandlockouts  

\usepackage{amsfonts,amssymb,amsmath,mathtools,multicol}
\usepackage{physics}
\usepackage{multirow}
\usepackage{float}
\usepackage{graphicx}
\graphicspath{{images/}}

\usepackage{cite}
\usepackage{hyperref}
\usepackage{subcaption}
\usepackage[dvipsnames]{xcolor}

\usepackage{algorithm}
\usepackage[indLines=true, 
		     noEnd=true, 
		     beginLComment=//~, 
		     endLComment=~, 
		     beginComment=//~]{algpseudocodex}
\renewcommand{\algorithmicrequire}{\textbf{Input:}}
\renewcommand{\algorithmicensure}{\textbf{Output:}}

\usepackage[inline]{enumitem}
\usepackage{cancel}

\usepackage{tabularx}   
\usepackage{booktabs}   
\usepackage{makecell}

\newcommand{\HL}[1]{{\color{black}{#1}}}

\newcommand{\V}[1]{{\boldsymbol{\mathbf{#1}}}}

\newcommand{\Vdot}[1]{\dot{\V{#1}}}

\newcommand{\R}{\mathbb{R}}
\newcommand{\rarr}{\rightarrow}

\newcommand{\T}{\top}

\newcommand{\Def}{\coloneqq}

\newcommand{\bmat}[1]{\bmqty{#1}}
\newcommand{\MAT}[1]{\bmat{#1}} 
\usepackage[normalem]{ulem}
\title{Planning for quasi-static manipulation tasks via an intrinsic haptic metric\HL{: a book insertion case study}}
\author{Lin Yang, Sri Harsha Turlapati, Chen Lv, Domenico Campolo$^*$
\thanks{All authors are with the School of Mechanical and Aerospace Engineering, Nanyang Technological University (NTU), Singapore.}
\thanks{$^*$ Corresponding author: {\tt d.campolo@ntu.edu.sg}}
\thanks{This research is supported by the National Research Foundation, Singapore, under the NRF Medium Sized Centre scheme (CARTIN).}
}
\date{}

\begin{document}
\maketitle

\begin{abstract}

Contact-rich manipulation often requires strategic interactions with objects, such as pushing to accomplish specific tasks.
We propose a novel scenario where a robot inserts a book into a crowded shelf by pushing aside neighboring books to create space before slotting the new book into place. Classical planning algorithms fail in this context due to limited space and their tendency to avoid contact. Additionally, they do not handle indirectly manipulable objects or consider force interactions. 
Our key contributions are: $i)$ reframing quasi-static manipulation as a planning problem on an implicit manifold derived from equilibrium conditions; $ii)$ utilizing an intrinsic haptic metric instead of ad-hoc cost functions; and $iii)$ proposing an adaptive algorithm that simultaneously updates robot states, object positions, contact points, and haptic distances.
We evaluate our method on a crowded bookshelf insertion task, and it can be generally applied to rigid body manipulation tasks.
We propose proxies to capture contact points and forces, with superellipses to represent objects. This simplified model guarantees differentiability. 
Our framework autonomously discovers strategic wedging-in policies while our simplified contact model achieves behavior similar to real world scenarios. We also vary the stiffness and initial positions to analyze our framework comprehensively. The video can be found at \url{https://youtu.be/eab8umZ3AQ0}.

\end{abstract}



\section{Introduction}

\begin{figure}[H]
    \centering
    \begin{subfigure}[b]{0.37\linewidth} 
        \centering
        \includegraphics[width=\linewidth]{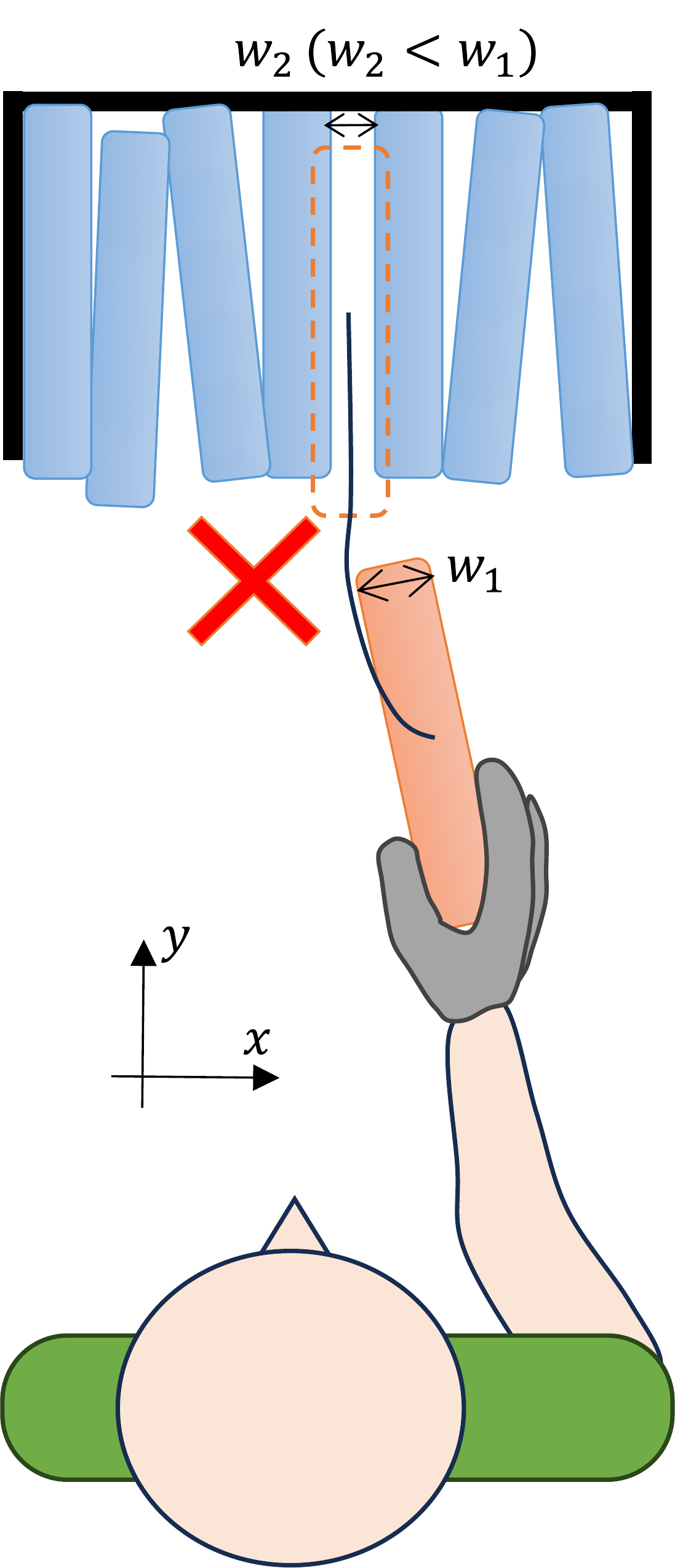}
        \caption{A traditional planner fails to place the book due to limited space.}\label{fig:shelf}
    \end{subfigure}\hspace{2mm} 
    \begin{subfigure}[b]{0.3\linewidth} 
        \centering
        \begin{subfigure}[b]{\linewidth}
            \centering
            \includegraphics[width=0.75\linewidth]{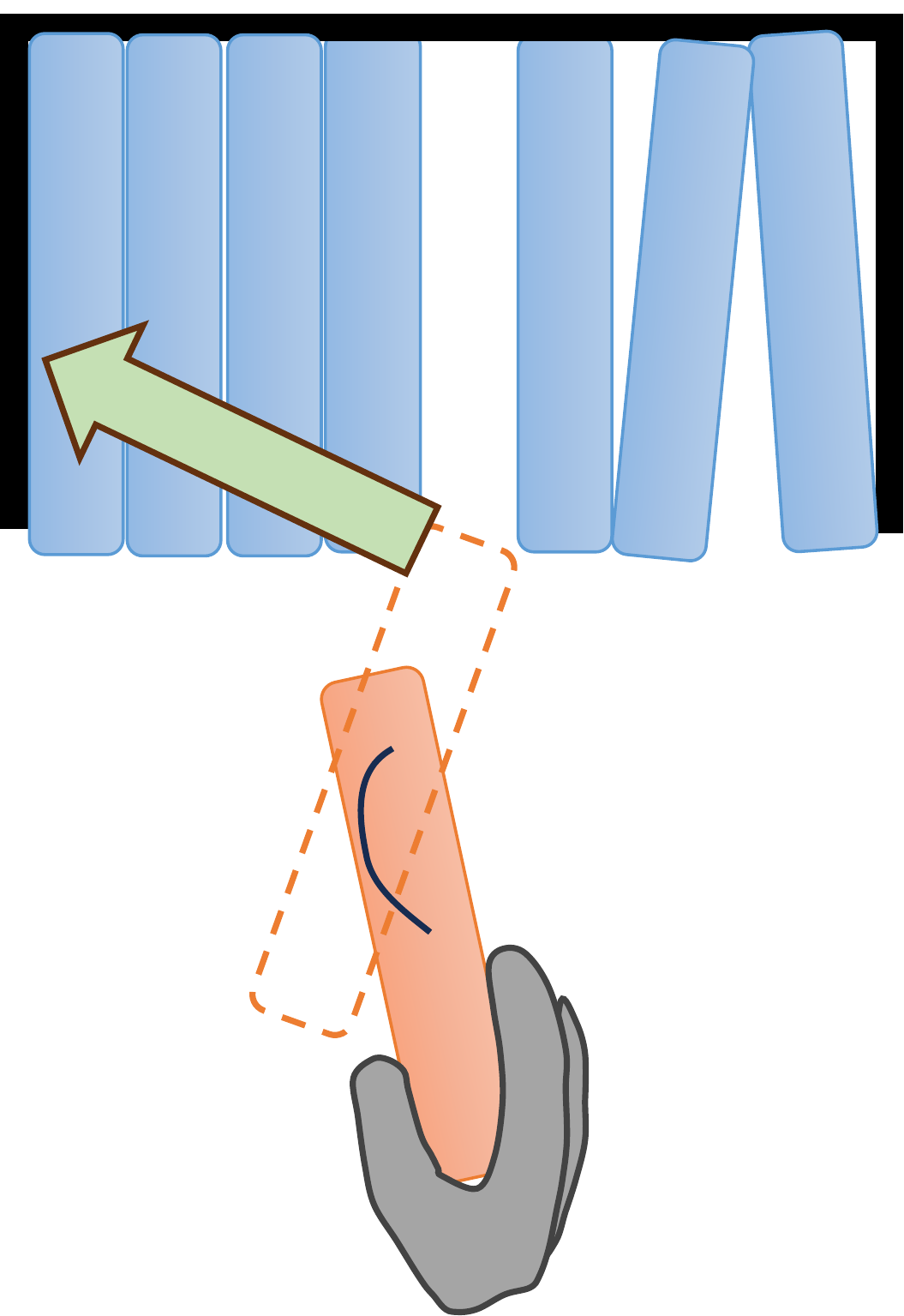} 
            \caption{The person must first push neighboring books aside to create space.}\label{fig:intro:push}
        \end{subfigure}
        \medskip
        \begin{subfigure}[b]{\linewidth}
            \centering
            \includegraphics[width=0.75\linewidth]{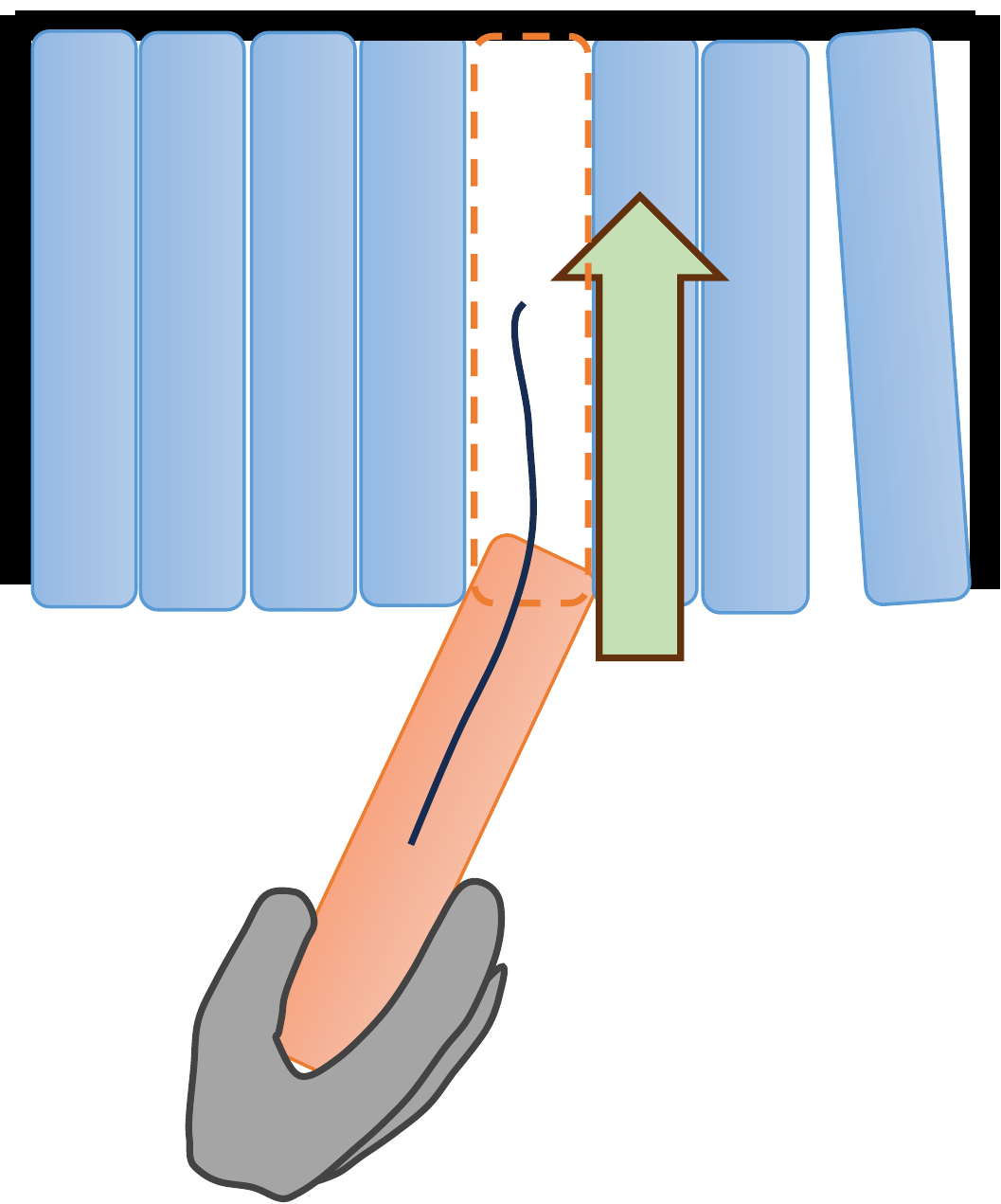} 
            \caption{The person then applies a forward force to slide the book into the shelf.}\label{fig:intro:insert}
        \end{subfigure}
    \end{subfigure}
    \caption{Inserting a book into a crowded shelf filled with books (blue), where the space is insufficient for a new book (orange).}\label{fig:1}
\end{figure}

Robotic manipulation is a challenging area that not only requires conventional motion planning but also the ability to maintain and control physical contact with objects \cite{suomalainen2022survey}. 
Unlike traditional motion planning focusing on avoiding collisions in free spaces, manipulation planning must account for contact and strategic force interaction.
As shown in Fig. \ref{fig:1}, inserting a book into a crowded shelf typifies a contact-rich task, where robots must navigate tight spaces and manage precise contact interaction. 
Traditional planning algorithms \cite{lavalle2006planning} and constraint motion planning \cite{kingston2018sampling}, being solely based on geometry, invariably fail when the book to be inserted is wider than available slot in the bookshelf, underscoring the need for advanced frameworks that integrate motion and contact interaction.

Geometry alone is insufficient for such contact-rich tasks. Incorporating haptics (i.e., contact forces) is essential for robots to effectively handle contact-rich tasks. A common approach to address contact forces involves quasi-static assumptions, which can be useful for tasks like the one described. 
Quasi-static manipulation simplifies the problem by focusing on contact forces while neglecting inertial and Coriolis effects \cite{pang2023planning}. This assumption has been widely applied in tasks such as assembly \cite{whitney1982quasi} and manipulation \cite{ozawa2017grasp}. However, traditional quasi-static approaches often face several limitations. These methods typically rely on (i) predefined compliant structures to handle contact dynamics, (ii) precise geometric information about objects and environments, and (iii) user-defined contact phases to manage transitions during interaction \cite{salem2020robotic, davchev2022residual,ozawa2017grasp,katayarna2022quasistatic}.
\vspace{-3mm}
\vspace{-3mm}
While quasi-static manipulation is one approach for contact-rich tasks, it heavily relies on user-defined contact phases to manage interaction transitions. Reinforcement learning (RL) has been explored as alternatives to address this limitation \cite{elguea2023review}. 
Some model-free algorithms are capable of handling contact-rich tasks after extensive exploration, however, RL often relies on task-specific cost functions \cite{fan2024learning, yang2023planning}, long training time \cite{azulay2022haptic} and suffer from sim2real gap, reducing its generalization to real-world tasks \cite{elguea2023review}.
The combination of dynamic movement primitives (DMP) and black-box optimization (BBO) \cite{stulp2013robot} simplifies the learning problem by enabling efficient exploration in a low-dimensional parameter space, reducing the required training time.
Meanwhile, impedance control mitigates sim2real gap in some extent \cite{chen2023multimodality}.
Furthermore, differentiability is important for optimization since the gradient offers advantages in task inversion  \cite{toussaint2018differentiable}.
These methods highlight the need for a unified approach that eliminates manual phase definitions, avoids specialized cost designs, and models contact interactions.

\begin{table*}
\caption{\HL{Representative prior work on book manipulation}}
\label{tab:robotic-books}
\begin{small}
\begin{tabular}{|p{2.2cm}|p{2.8cm}|p{3.2cm}|p{2.8cm}|p{4.2cm}|}
\hline
\textbf{Reference} & \textbf{Task} & \textbf{Approach} & \textbf{Robot/Control} & \textbf{Remarks} \\
\hline
\multicolumn{5}{c}{\textbf{Book insertion}} \\
\hline
[Sygo23] \cite{sygo2023multi} & Rearranging books on a shelf & Multi-stage perception + collision-aware manipulation & PR2 + human-like robot hand & Collision-aware planned motion + force-controlled placement policy to push book sideways \\
\hline
[Nakajima11] \cite{nakajima2011study} & Insert book into narrow gap on bookshelf & Grasp, incline, and push book aside & Parallel-jaw gripper with position/force control & Friction modeling between book and shelf is key \\
\hline
\multicolumn{5}{c}{\textbf{Book extraction}} \\
\hline
[Comsa14] \cite{comsa2014automated}, [Modler14] \cite{modler2014dedicated} & Extract books from shelf & Book handling gripper mechanism & Mobile platform + linear drives + robotic arm & Learned book positions \\
\hline
[Ikeda24] \cite{ikeda2024retrieving} & Extract binder from shelf & Claw to grasp binder & 6DOF foldable hand / Pseudo-curved trajectory & Strategy to tilt and grasp a binder \\
\hline
[Prats04] \cite{prats2004autonomous}, [Prats06] \cite{prats2008uji}, [Prats07] \cite{prats2007towards} & Book extraction & Vision-based detection; force feedback to fit gripper width & UJI Librarian robot / Position + force control & Barrett hand pushes book to tilt then grasps from side \\
\hline
[Heyer12] \cite{heyer2012book} & Book detection & Autonomous detection, grasp and place & Assistant robot / Collision-aware path planning & Discusses tolerances in crowded bookshelves (Sec 4.1) \\
\hline
\end{tabular}
\end{small}
\end{table*}

To address the limitations of existing approaches, we build upon a novel framework for quasi-static manipulation introduced in \cite{campolo2025geometric, campolo2023quasi}, where the system is modeled using a manipulation potential. This potential combines robot impedance control \cite{siciliano2008springer} and physical contact modeling, enabling manipulation tasks to be formulated as an optimization problem.
\HL{In contrast to methods that require explicitly defined contact points \cite{doshi2022manipulation, pang2023planning}, our framework computes them implicitly within the manipulation potential.} The framework splits system variables into internal states $\V z$ and control inputs $\V u$, where the inputs $\V u$ guide the movement of uncontrollable states $\V z$ along an implicitly defined equilibrium manifold ($\mathcal M^{eq}$). To evaluate and optimize the manipulation process, \cite{campolo2025geometric, campolo2023quasi} introduces an intrinsic haptic metric that quantifies the haptic distance between equilibrium states, providing a task-independent measurement.

We extend this framework by introducing a haptic planning algorithm, characterized by a manipulation potential. Our method proposes an adaptive algorithm that continuously computes: (i) the haptic distance using the intrinsic haptic metric, and (ii) the positions of the robot, uncontrollable objects, and contact points. 
To ensure the framework’s differentiability, we employ superellipse-based object representations \cite{jaklic2000segmentation} and proxies \cite{kana2021human} to accurately capture contact points. Unlike classical quasi-static approaches and motion planning algorithms, our method combines (a) impedance controlled manipulation, (b) supports arbitrary object geometries, and (c) the automatic evolution of contact phases based on equilibrium states, without relying on extensive pre-definition. The intrinsic nature of our framework allows it to continuously explore and optimize along the equilibrium manifold, seamlessly integrating physical interaction with robot impedance control.

We validate the framework with a challenging crowded book insertion task. Previous studies on robot librarians mainly focus on book extraction \cite{ikeda2024retrieving}, often assuming sufficient gaps for gripper insertion. In contrast, book insertion involves challenges: (i) approaching narrow gaps, (ii) deciding pushing direction and force, and (iii) determining action termination. Existing methods \cite{nakajima2011study, sygo2023multi} rely on pre-designed, hierarchical policies tailored to specific tasks. \HL{For further comparsion, please refer to Table \ref{tab:robotic-books}}. Our framework continuously computes optimal policies based on the bookshelf model, offering key advantages: (i) the inclined approach is generated automatically, 
(ii) haptic distance with haptic metric determines force and pushing direction to overcome resistance, and 
(iii) the DMP termination condition identifies when to stop.

\vspace{-1mm}
\section{Methodology: Manipulation planning on equilibrium manifold}\label{sec: equilibrium mfd}

In this section, we detail our configuration space and the key components, e.g., haptic metric, haptic distance and adaptive Ordinary Differential Equations (ODE).

\subsection{Quasi-Static Mechanical Manipulation System}
Under the assumption of quasi-static motions, the mechanical robot-environment interaction is reformulated as a control problem via splitting of variables $\mathcal{Z} \times \mathcal{U}$ \cite{campolo2025geometric,campolo2023quasi}, where $\V z\in\mathcal{Z}\subset\R^{N}$ is the \textit{internal state} (also referred to as uncontrollable objects) and $\V u\in\mathcal{U}\subset\R^{K}$ is the \textit{control} of the robot (which can be interpreted as the desired pose in impedance control).
The configuration of the system is determined solely by its potential energies, such as elastic and gravitational energies, which we refer to as the manipulation potential $W(\V z, \V u)$.
Define manipulation potential as a smooth field on the space $W: \mathcal{Z}\times\mathcal{U} \rarr \R$. For any given control $\V u$, equilibria of system satisfy
\begin{equation}
\partial_{\V z}W(\V z^*(\V u), \V u) = \V 0 \in\R^{N}.
\label{eq: W_z}
\end{equation}
We define  $\partial_{\V q} W \equiv [\partial_{q_1} W, \ldots, \partial_{q_a} W]^T$, 
where the \textit{partial} (column) operator is defined as $\partial_{\V q} = [\partial_{q_1}, \ldots, \partial_{q_a}]^T$. Meanwhile, define the shorthand notation $\partial_{\V z}^2 \equiv (\partial_{\V z} \partial_{\V z}^T) = \partial_{\V z} \partial_{\V z}^T$ for Hessians and mixed-derivative operators. Hence, in Eq. \ref{eq: W_z}, $\partial_{\V z}$ denotes the gradient with respect to $\V z$, which means internal forces acting on objects $\V z$. Under the quasi-static assumption, the total force acting on the objects should be zero.

A point is strictly stable when its Hessian is positive definite, i.e., $\partial^2_\V{zz}W|_{\ast} \succ 0$. Assuming the Hessian $\partial^2_\V{zz}W\in\R^{N\times N}$ is of full rank when $\partial_{\V z}W(\V z^*, \V u) = \V 0$, via the \textit{implicit function theorem} \cite{spivak2018calculus}, the set
\begin{equation}
\mathcal{M}^{eq}\Def \{(\V z,\V u)\in\mathcal{Z}\times\mathcal{U} | \partial_{\V z}W(\V z,\V u)=\V 0\}
\label{eq:manifold}
\end{equation}
is a smooth embedded submanifold in the ambient space $(\mathcal{Z} \times \mathcal{U})$. We refer to $\mathcal{M}^{eq}$ as the \textit{equilibrium manifold} of the system. 
Under the quasi-static assumption, the external force equals the control force. Thus, from the robot’s perspective, it can use its joint torque sensors to estimate the external contact force. We define the control forces as
 $\V f_\text{ctrl} = -\partial_\V{u}W$ is called \textit{control forces} \cite{campolo2025geometric}.

\subsection{Haptic Metric and Haptic Distance}\label{sec: haptic metric}


The closeness of states is determined by a distance function. 
In the context of quasi-static manipulations, we choose the \textit{haptic metric} as the Riemannian metric of the control space $\mathcal{U}$. Following \cite{campolo2025geometric} the \textit{control Hessian} of an interconnected system with a potential $W: \mathcal{Z}\times\mathcal{U}\rarr\R$ is defined as:
\begin{equation}\label{eq: control Hessian}
\V G_m(\V z^*(\V u), \V u) \Def \partial^2_\V{uu}W - \partial^2_\V{uz}W (\partial^2_{\V{zz}}W)^{-1} \partial^2_\V{zu}W,
\end{equation}
This is computed as the Schur's complement of the Hessian of the potential $W(\V z^*_m, \V u)$, evaluated at equilibrium (i.e., $\V z^*(\V u)$ s.t. $\partial_\V{z}W(\V z^*, \V u)=\V 0$), and
$\partial^2_\V{uz}W = (\partial^2_\V{zu}W)^\T \in \R^{K\times N}$. The squared Hessian $\V G^2_m(\V z^*(\V u), \V u)$ is called the \textit{haptic metric}, which is at least positive semi-definite. 
Following \cite{campolo2025geometric}, for control policy $\V u: [0,1]\rarr\R^K$ connecting two points on the control space, haptic distance $S$ between any two points $\V u(0)$ to $\V u(1)$ is defined as,
\begin{equation}\label{eq: geodesic U}
S[\V u] = \int_{0}^{1} \sqrt{\dot{\V {u}}^{T} \V G^{2}_m(\V z^*(\V u), \V u) \dot{\V {u}}} \; dt
\end{equation}
The greater the force required by robot during manipulation, the larger the value of $S$. 
However, since $\V G_m(\V z^*(\V u), \V u)$ depends on $\V z^*(\V u)$, and the equilibrium manifold is defined implicitly, the value of $\V z^*(\V u)$ remains unknown. The next step, therefore, is to determine a method for computing $\V z^*(\V u)$.



\subsection{Exploring implicit manifold via adaptive ODE}
\begin{figure}[!h]
\centering
\includegraphics[width=0.45\textwidth]{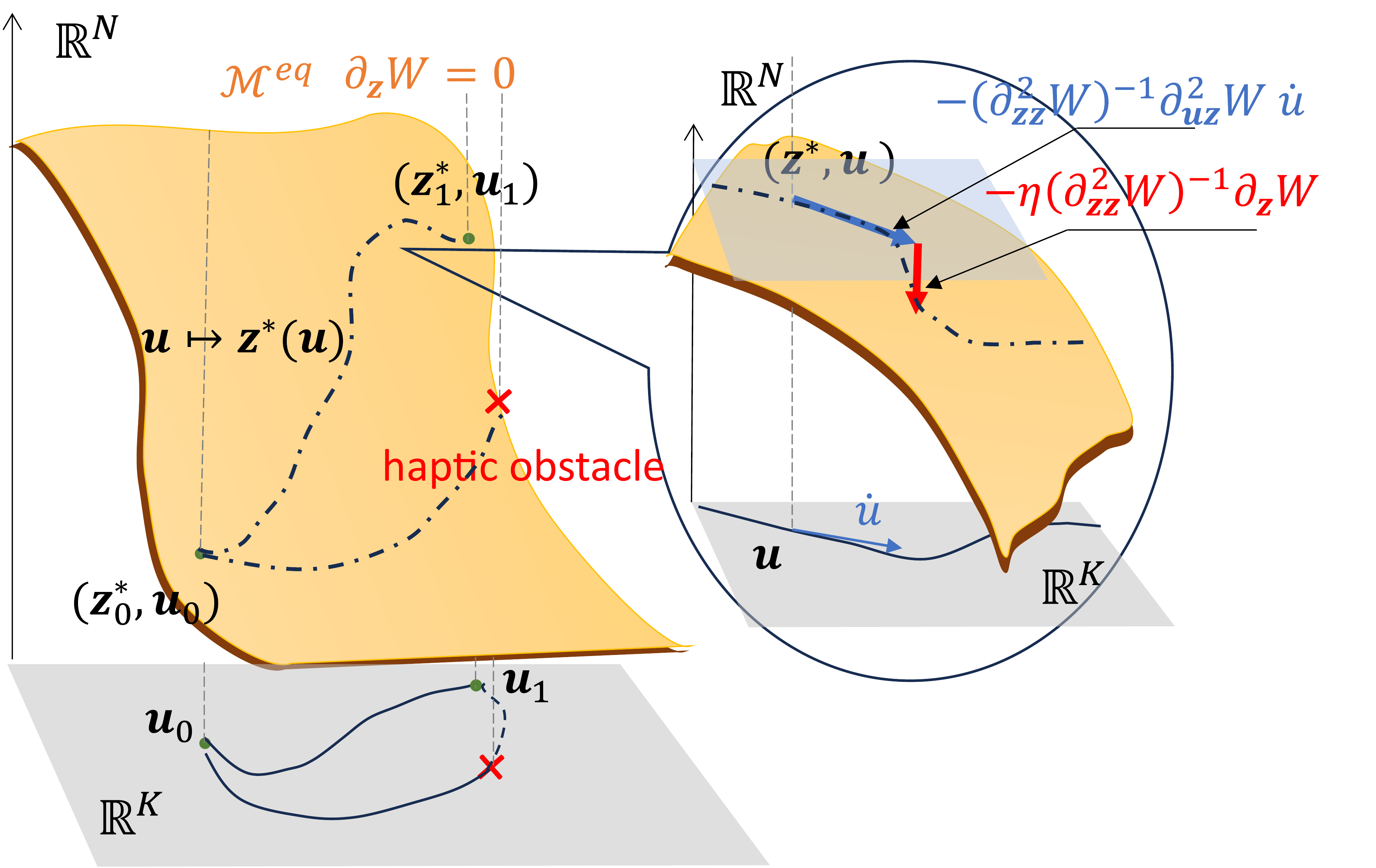}
\caption{Adaptive ODE allows to lift control $\V u(t)$ onto $\mathcal{M}^{eq}$.
$\V u$ is extended by moving along $\Vdot u$ and updating $\V z$ on the $\mathcal{M}^{eq}$ via Eq. \ref{eq:adaptive ODE}. Blue arrow denotes $\V z$ linear approximation as the variation of $\V u$, red arrow represents Newton-Raphson ‘infinitesimal’ adjustment. Policy can be terminated by haptic obstacle.}
\label{fig: adaptive ODE}
\end{figure}

To compute $\V z^*(\V u)$ online, we propose an adaptive algorithm to explore the implicit manifold.
This method initiates from a point $(\V z^*_0, \V u_0)$ to update the state $\V z$ in the equilibrium manifold as the control $\V u$ moves along the direction $\Vdot u$. A first order approximation on $\mathcal{M}^{eq}$ is given as,
\begin{equation}
\Vdot z = -(\partial^2_\V{zz}W)^{-1}\partial^2_\V{uz}W \Vdot u
\label{eq:dzdu}
\end{equation}
Equation \ref{eq:dzdu} is depicted as blue arrow in Fig. \ref{fig: adaptive ODE}, illustrating the linear relationship between the infinitesimal changes in $\V z$ and $\V u$. Computing an implicit manifold from a given set of (nonlinear) equations typically relies on standard iterative methods, such as the Newton-Raphson method, and
\cite{schneebeli2011newton} has shown classical Newton-Raphson method can be reformulated as an adaptive ODE.
Based on quasi-static assumptions, we propose tracking the evolution $t \to \V z(t) \in \R^N$ as control parameters $t \to \V u(t) \in \R^K$ by numerically solving a set of adaptive ODE. The Newton-Raphson 'infinitesimal' adjustments (Eq. \ref{eq:dz}), when $\V u$ is held constant, lead to out-of-equilibrium dynamics, represented by the red arrow in Fig. \ref{fig: adaptive ODE}.
\begin{equation}
\Vdot z = - \eta(\partial^2_\V{zz}W)^{-1} \partial_\V z W,
\label{eq:dz}
\end{equation}
where $\eta$ represents step size in the ODE \cite{schneebeli2011newton}, which influences the convergence of the ODE.
If $\eta$ is too small, the ODE may fail to converge within the given time. Conversely, if $\eta$ is too large, the ODE might jump to a different branch of $\mathcal M^{eq}$.
Combining both behaviors (Eq. \ref{eq:dz}) and (Eq. \ref{eq:dzdu}) we can define the compound dynamics as $\Vdot z$, 
\begin{equation}
\Vdot z = -(\partial^2_\V{zz}W)^{-1}\partial^2_\V{uz}W \Vdot u - \eta(\partial^2_\V{zz}W)^{-1} \partial_\V z W \label{eq:adaptive ODE}
\end{equation}
which is represented as the dot-line curve on $\mathcal M^{eq}$ in Fig. \ref{fig: adaptive ODE}. To sum up, this method allows to lift a curve $\V u(t)$ on the control space onto the equilibrium manifold, plotted as dot-line curve.

\subsection{Haptic Obstacle}
In manipulation tasks, certain obstacles cannot be traversed by the robot. Within our configuration space, we require the node must always remain on the $\mathcal M^{eq}$. Therefore, we define a haptic obstacle as follows:
\begin{equation}\label{eq: Haptic Obstacle}
\rm{det} ( \partial_{\V z \V z} W (\V z^*(\V u), \V u)) \geq \lambda > 0
\end{equation}
Equation \ref{eq: Haptic Obstacle} consists of two components.
The term $\lambda>0$ indicates Hessian is positive definite, suggesting the node is stable on $\mathcal{M}^{eq}$ \cite{campolo2025geometric}. 
The condition $\rm{det} ( \partial_{\V z \V z} W (\V z, \V u)) \geq \lambda$ is imposed because the inverse of Hessian is required in Eq. \ref{eq:adaptive ODE}. 
$(\partial^2_{\V z \V z}W)^{-1}$ should be sufficiently small to avoid $\V z$ change significantly. 
Meanwhile, stable node allows us to apply the implicit function theorem to derive Eq. \ref{eq:dzdu}, with such an equilibrium referred to as non-critical \cite{campolo2025geometric}. This ensures stability and continuity in the manipulation process while avoiding singularities.


\section{Application of manipulation planning: Crowded Bookshelf Insertion} \label{sec: book model}

In this section, we introduce bookshelf model, which incorporates superellipse representation, proxies, and an augmented ODE. 
Traditional motion planners would fail in this context due to the limited space available for a new book. In contrast, our framework effectively handles uncontrollable objects, captures contact interactions, and employs a haptic metric to evaluate the optimality. 
This method enables us to tackle the challenges inherent in the crowded bookshelf insertion.

\subsection{Superellipses and proxy}

We ensure differentiability of the manipulation potential by representing books with superellipses and using proxies to capture contact points. \HL{SQ can fit common object shapes from point cloud data \cite{yang2024path}.}

\subsubsection{Superellipses}
Since books are typically rectangular, we use superellipses (SQ) to represent them. A SQ is implicitly defined by the equation:
\begin{align}
    \left(\frac{x}{a_1}\right)^{\frac{2}{\epsilon}} + \left(\frac{y}{a_2}\right)^{\frac{2}{\epsilon}} = 1
\end{align}
Here, $\epsilon$ controls the shape of the SQ, while $a_1$ and $a_2$ define its size.
Additionally, define a inside-outside function $F(x,y)$ as 
\begin{align}
    F(x,y) &= \left(\frac{x}{a_1}\right)^{\frac{2}{\epsilon}} + \left(\frac{y}{a_2}\right)^{\frac{2}{\epsilon}} - 1
    \label{eq:io}
\end{align}
The function $F(x,y)$ possesses a useful property. For a point $(x_0,y_0)$, if $F(x_0,y_0) > 0$,the point is outside the SQ. If $F(x_0,y_0) = 0$, the point lies on the surface, and if $F(x_0,y_0) < 0$, the point is inside. Due to this property, the function F is also referred to as the \textbf{inside-outside} function.
Analogous to a circle, a SQ can be parameterized by angular parameterization $\gamma$ as follows \cite{jaklic2000segmentation},
\begin{align}
    \V p (\gamma) & =  \V r(\gamma) \begin{bmatrix}
    \cos \gamma \\
    \sin \gamma
    \end{bmatrix}, 0 \leq \gamma \leq 2\pi \nonumber 
    \\
        \V r(\gamma) &= \frac{1}{\sqrt{\left( \frac{\cos \gamma}{a_1} \right)^{\frac{2}{\epsilon}} + \left( \frac{\sin \gamma}{a_2} \right)^{\frac{2}{\epsilon}}}}.
        \label{2.72}
\end{align}

\subsubsection{Proxy}
\begin{figure}[!h]
  \centering
  \begin{subfigure}{0.25\textwidth}
    \includegraphics[width=\linewidth]{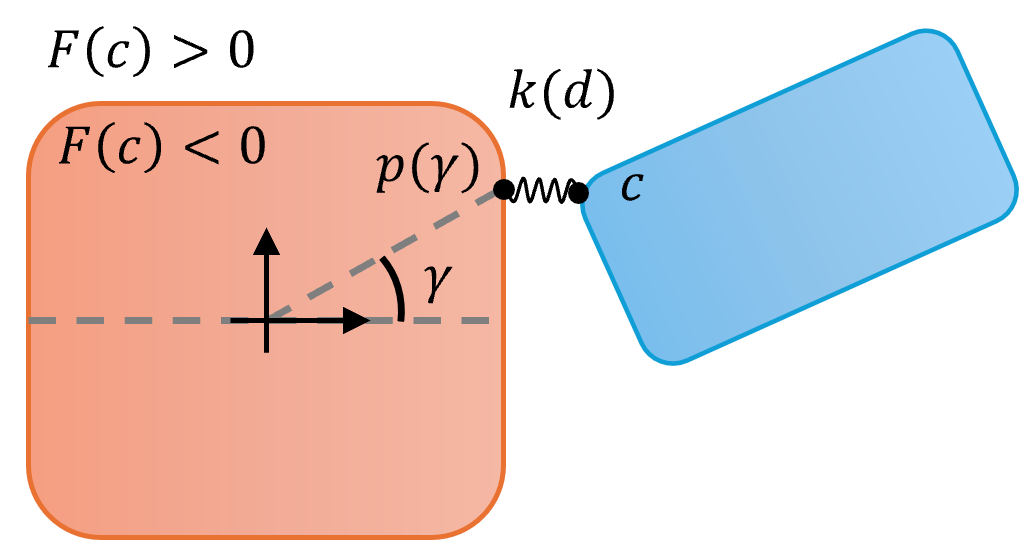}
    \caption[]
    {\small Proxy on superellipse (before contact).}
    \label{fig:SQ_proxy}
  \end{subfigure}
  \hspace{2mm} 
  \begin{subfigure}{0.2\textwidth}
    \includegraphics[width=\linewidth]{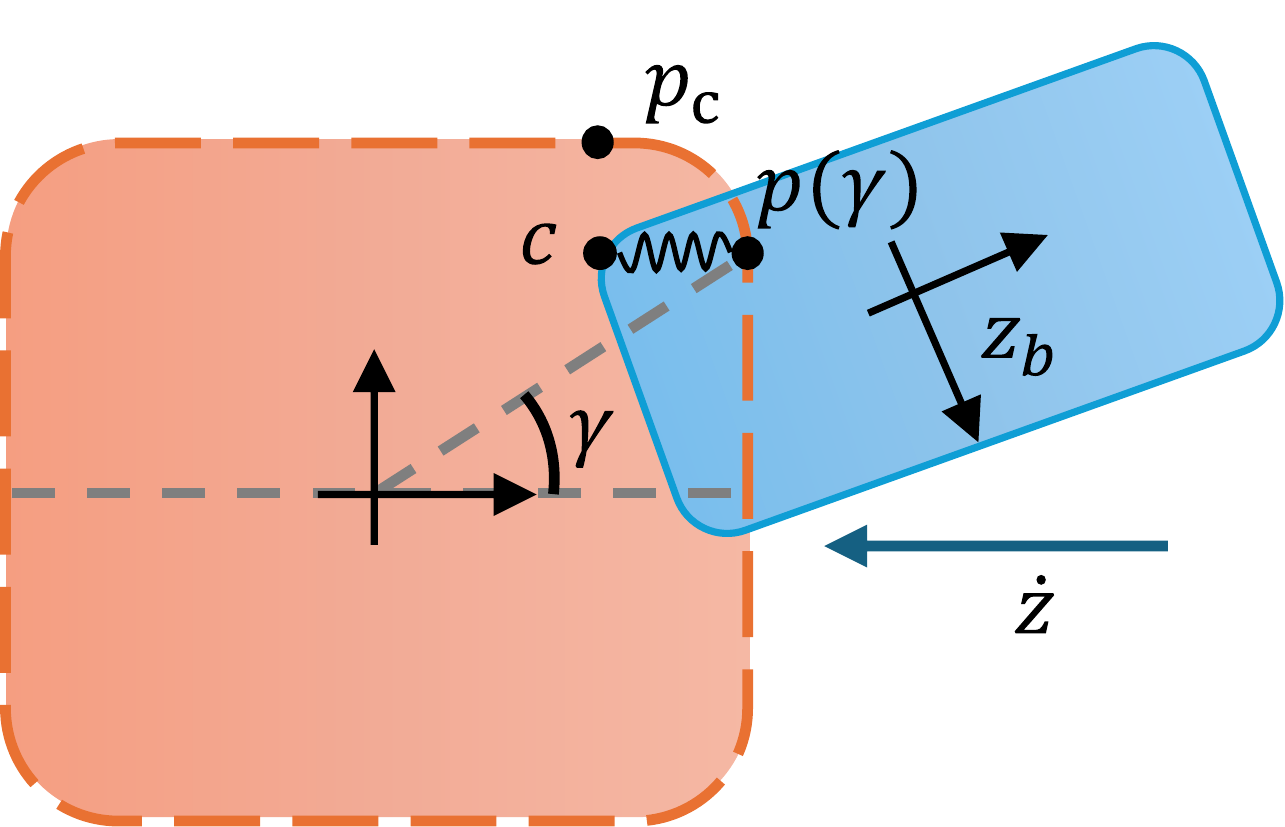}
    \caption[] 
    {\small \HL{Proxy on superellipse (after contact).}}
    \label{fig:SQ_proxy_after_ct}
  \end{subfigure}
  \caption{\HL{Behavior of proxy on superellipse.}} 
  \label{fig:proxy}
\end{figure}

We utilize variable $\gamma$ to represent the contact point $\V p(\gamma) \in \R^2$ on the book, referred to as \textit{proxy}.
This allows us to represent contact interactions using both the proxy and the inside-outside function.
Define a book $\V z_b = [z_x, z_y, z_\theta]^T \in SE(2)$, symbolize one corner of a book $\V z_b$ as $\V c(\V z)$ with a corresponding proxy  on another book. The proxy represents the contact point, so it should be the closest point to the corner. This condition can be expressed as:
\begin{align}
    \underset{\gamma }{\arg \min } \; &: \norm{\V c(\V z) - \V p(\gamma)} , \; \gamma \in [0, 2\pi] \label{eq:proxy}
\end{align}
Meanwhile, the nonlinear stiffness $k(d)$ between contact points can be modeled using the signed distance $d= F(\V c)$ in the book frame.
\begin{equation}\label{eq: stiffness}
k(d) = k_\text{min} + \frac{1-\tanh(d/d_0)}{2} k_\text{max}.
\end{equation}
Here, $d$ regularizes mechanical contact.
If $d<0$, it indicates contact between the two books, leading to a large contact force due to the high stiffness $k_{max}$.
Conversely, when there is no contact ($d>0$), the contact force should be negligible, reflected by a significantly lower stiffness ($k_{max} \gg k_{min}$). Based on the penetration depth $d_0$, the stiffness can be modulated to reflect the physical properties of the material, making it a powerful model for simulating contact interactions.
The idea of proxy and SQ is illustrated in Fig. \ref{fig:proxy}, SQ provides a practical and effective method for modeling a book, ensuring smoothness for gradient-based computations. This parametrization $\gamma$, referred to as a proxy, is designed to continuously capture the contact point \HL{while avoiding discontinuities \cite{le2024contact}, even during interpenetration between objects. To specify, the proxy remains at the closest point on the SQ before contact occurs, as illustrated in Fig. \ref{fig:SQ_proxy}. After contact, as the SQ penetrates the object, the proxy gradually moves toward the actual closest point $\V p_c$, as shown in Fig. \ref{fig:SQ_proxy_after_ct}.}

\vspace{-2mm}


\subsection{Crowded bookshelf model}
\begin{figure}[!h]
\centering
\includegraphics[width=0.7\linewidth]{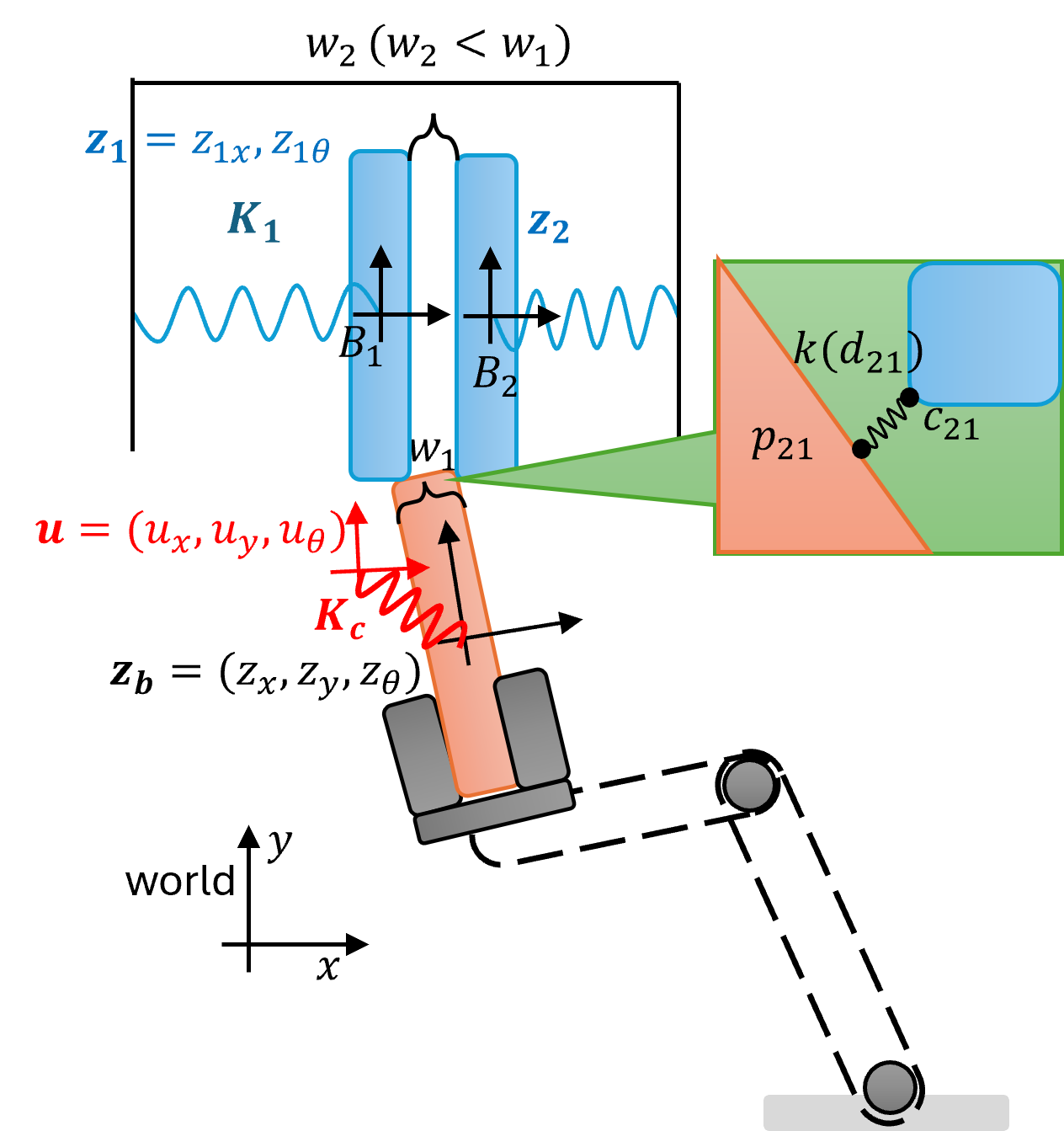}
\caption{A book need to be inserted to a narrow shelf, where the remaining space is not enough for insertion. The manipulated book $\V z_b$ is controlled with impedance control policy $\V u$. Contact interaction is captured by proxy, and the resistance among the books on the bookshelf is captured by external stiffness $\V K_i$.}
\label{fig:bookmodel}
\end{figure}

As depicted in Fig. \ref{fig:bookmodel}, a robot manipulates a book within a 2D space, defined by $\V u =[u_x, u_y, u_\theta]^T \in SE(2)$. The gripper, employing impedance control, grasps the book using a diagonal stiffness matrix $\V K_c$.
The book in the robot's gripper is described by $\V z_b=[z_x, z_y, z_\theta]^T \in SE(2)$, while other two books on the shelf are represented by $\V z_1, \V z_2$, each possessing only 2 DOF allowing horizontal movement and rotating ($z_{ix}, z_{i\theta}, \; i = 1,2$). The resistive properties of the books are modeled as springs with a constant diagonal stiffness matrix $\V K_i = \text{diag}(k_{ix}, k_{i\theta})$. The initial position (equilibrium position) is labeled as $\V z_{i,0}$.
Moreover, the contact between books is captured by proxy. We denote corner as $\V c_{ij}$, and the position of proxy as $\V p_{ij}(\gamma_j)$, where $i$ denotes the index of object and $j$ denotes the index of corner. For each corner $j$, there is a corresponding proxy parameterized by $\gamma_j$. The number of proxies $N_p$ equals the number of corners. Symbolize as all proxy variable as $\V \Gamma \in \mathbb{T}^{N_p}$.
\vspace{-3mm}

Ultimately, we consider four proxy pairs ($N_p$ = 4): one corner on book $\V z_1$, one corner on book $\V z_2$, two corners on book $\V z_b$. Consequently, the configuration is defined as $\V z =[\V z_b, \V z_1, \V z_2, \V \Gamma]^T$. The manipulation potential equals,
\begin{align}\label{eq: W_tot_book}
{W}(\V z^*, \V u) & =  W_\text{ctrl}+W_\text{resist} +W_\text{contact}, \nonumber \\
&= \frac{1}{2}  (\V u - \V z_b)^T \V K_c (\V u - \V z_b) \nonumber \\
&+\sum_{i=1,2}  \frac{1}{2} (\V z_i - \V z_{i,0})^T \V K_i (\V z_i - \V z_{i,0})  \nonumber \\
&+ \sum_i \sum_j \frac{1}{2} k(d_{ij}) \|\V c_{ij}(\V z_i) - \V p_{ij}(\gamma_j) \|^2
\end{align}
This potential consists of three parts: $ W_\text{ctrl}$ denotes control energy, $W_\text{resist}$ accounts for the stiffness of the other books on the shelf. $W_\text{contact}$ captures the contact interaction between the manipulated book and the neighboring books.

\subsection{Augmented adaptive ODE}

We utilize Dynamic Motion Primitives (DMP) \cite{ijspeert2013dynamical} to represent our control $\V u(t)$, where $K$-dimensional controls (or \textit{policies}) $t\mapsto \V u_{\V \Theta}(t) \in\R^K$. 
DMP can also be viewed as a set of ODE, designed as nonlinear attractor systems that solve for a path from initial conditions to a final state.
Classical DMP can be used to map a finite dimensional set of parameters $\V \Theta \in\R^{K\times P}$ where $P$ is the number of Radial Basis Functions (RBFs) per degree-of-freedom (DOF) into smooth and differentiable functions.
\begin{equation}
    DMP:(\V \Theta, \V u_0, \V u_T, T)\mapsto (\V u_{\V \Theta}(t))
\end{equation}
while satisfying the boundary conditions $\V u_{\V\Theta}(0)=\V u_0$ and $\V u_{\V\Theta}(T)=\V u_T$, where $T$ represents the duration of the intended control input.
The initial condition of the robot $\V u_0$ is known from the robot state, while the final state $\V u_T$ is the empty space between neighboring books.
Thus, we combine all the ODEs as follows,
\begin{subequations}
\begin{align}
\tau\dot{\V u} &= \V v, \label{eq:dmp_1} \\
\tau\dot{\V v} &= \alpha_v(\beta_v(\V u_T - \V u) - \V v) + \V f(\V x), \label{eq:dmp_2} \\
\dot{\V z} &= \MAT{\mathbb{I}^{N-N_p} & \V 0 \\ \V 0 & \mathbb{I}^{N_p}}(-(\partial^2_{\V z \V z}W)^{-1} \partial^2_{\V u \V z}W \V v 
\nonumber \\
& - \eta (\partial^2_{\V z \V z}W)^{-1} \partial_{\V z}W - \alpha_p \partial_{\V z} {W}), \label{eq:dmp_3} \\
\dot{\phi} &= \sqrt{\V v^T \V G^2_m(\V u) \V v} \label{eq:dmp_5} 
\end{align}
\label{eq:allODE}
\end{subequations}
Equations \ref{eq:dmp_1} and \ref{eq:dmp_2} are the same in DMP, denoting the attractor system. Eq. \ref{eq:dmp_3} comes from adaptive ODE (Eq.\ref{eq:adaptive ODE}) and Eq.\ref{eq:proxy}, the first mask matrix separate objects and proxies, $\mathbb{I}^{N-N_p}$ multiply with adaptive ODE. The term with $\mathbb{I}^{N_p}$ (so the last term in bracket) computes the velocity of proxies, transferring a minimization problem (Eq.\ref{eq:proxy}) into an ODE. The (fastest) proxy dynamics are determined by the proxy damping constant $\alpha_p$.
The cost function $\phi$ (Eq.\ref{eq:dmp_5}) quantifies the effort required when the robot follows a control policy $\V u$. The greater the force required by the robot during manipulation, the larger the value of $\phi$, 
Meanwhile, we also setup geometric constraints for $\V z$, ensuring the exploration remains within the workspace.

\vspace{-2mm}
\begin{figure}[H]
\centering
\includegraphics[width=0.9\linewidth]{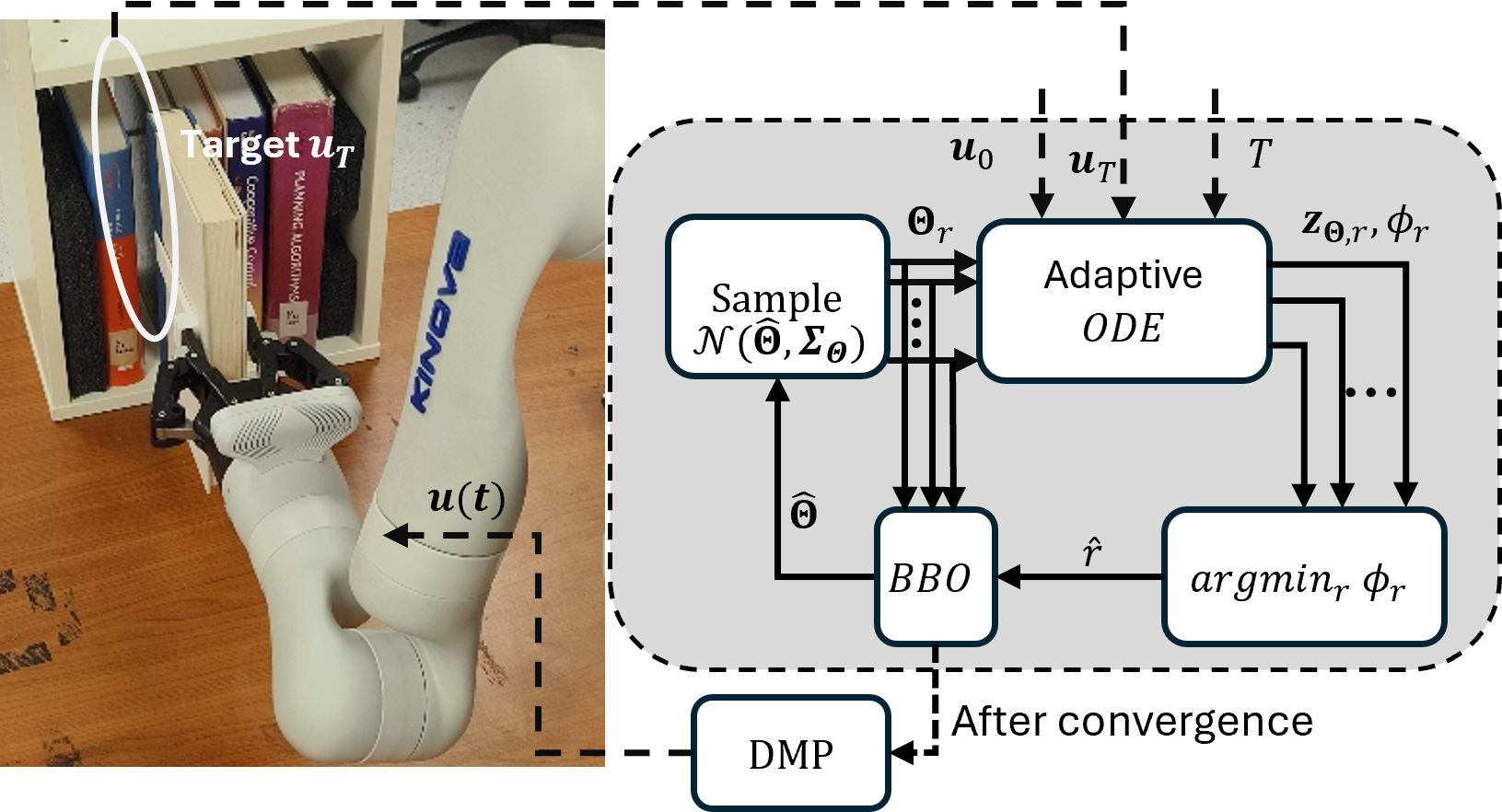}
\caption{
our BBO framework performs multiple rollouts with different DMP parameters, $\V \Theta^r$, calculating the corresponding cost (haptic distance) $\phi^r$ for each rollout. The optimal parameter, $\hat{\V \Theta}$, is then updated, and this process is repeated until the cost converges.}
\label{fig:BBO and Generalization}
\end{figure}
\subsection{Black-Box Optimization}
Following our previous work \cite{yang2023planning}, as illustrated in the grey block in Fig.\ref{fig:BBO and Generalization}. Given any initial position $\V u_0$, target $\V u_T$ and time duration $T$, we use BBO to find the optimal parameter $\hat{\V \Theta}$ for DMPs. In each iteration, we sample the parameter $\V \Theta_r \sim \mathcal{N}(\hat{\V \Theta}, \V \Sigma_{\Theta})$ based on optimal parameter in last iteration. After feeding these parameters into our augmented ODE (Eq. \ref{eq:allODE}), we compute corresponding $\V z_r$ and its cost $\phi_r$. Subsequently, the optimal parameter is updated.

After the cost converges, the optimal parameters are used to generate a control trajectory via DMP, specifically utilizing the first two rows in our augmented ODE (Eq. \ref{eq:allODE}). This control policy, $\V u(t)$, is then sent to the robot for real world implementation.
We choose an appropriate time duration to ensure that the inertial and velocity-dependent terms remain negligible compared to the contact forces.

\section{experiment validation}\label{sec: exp}

\begin{figure}[!h]
  \centering
  \begin{subfigure}{0.22\textwidth}
    \includegraphics[width=\linewidth]{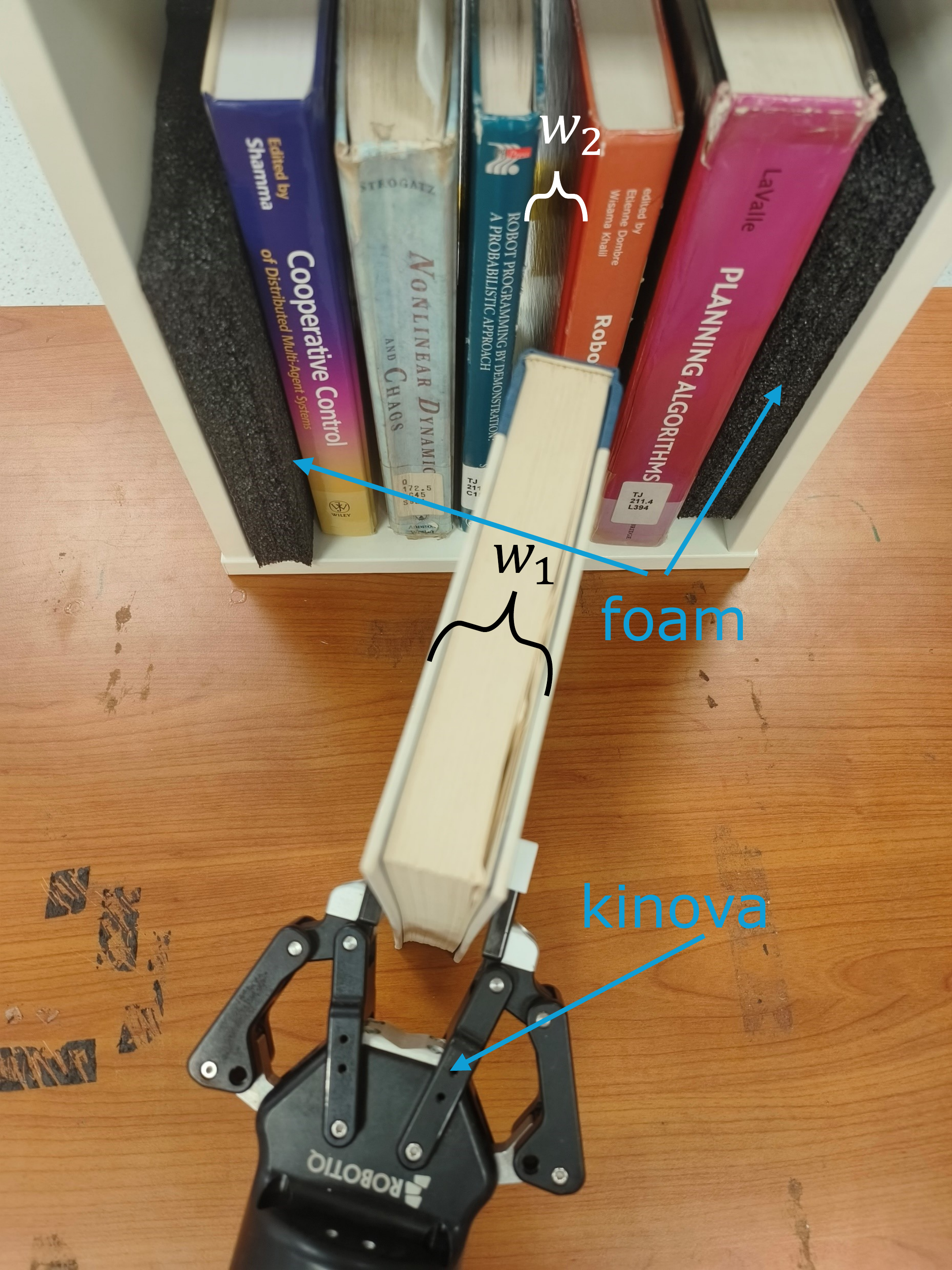}
    \caption[]
    {\small Experiment setup: The remaining space $w_2$ on the shelf is less than the width of the manipulated book $w_1$.}
    \label{fig:exp setup}
  \end{subfigure}
  \hspace{8mm} 
  \begin{subfigure}{0.185\textwidth}
    \includegraphics[width=\linewidth]{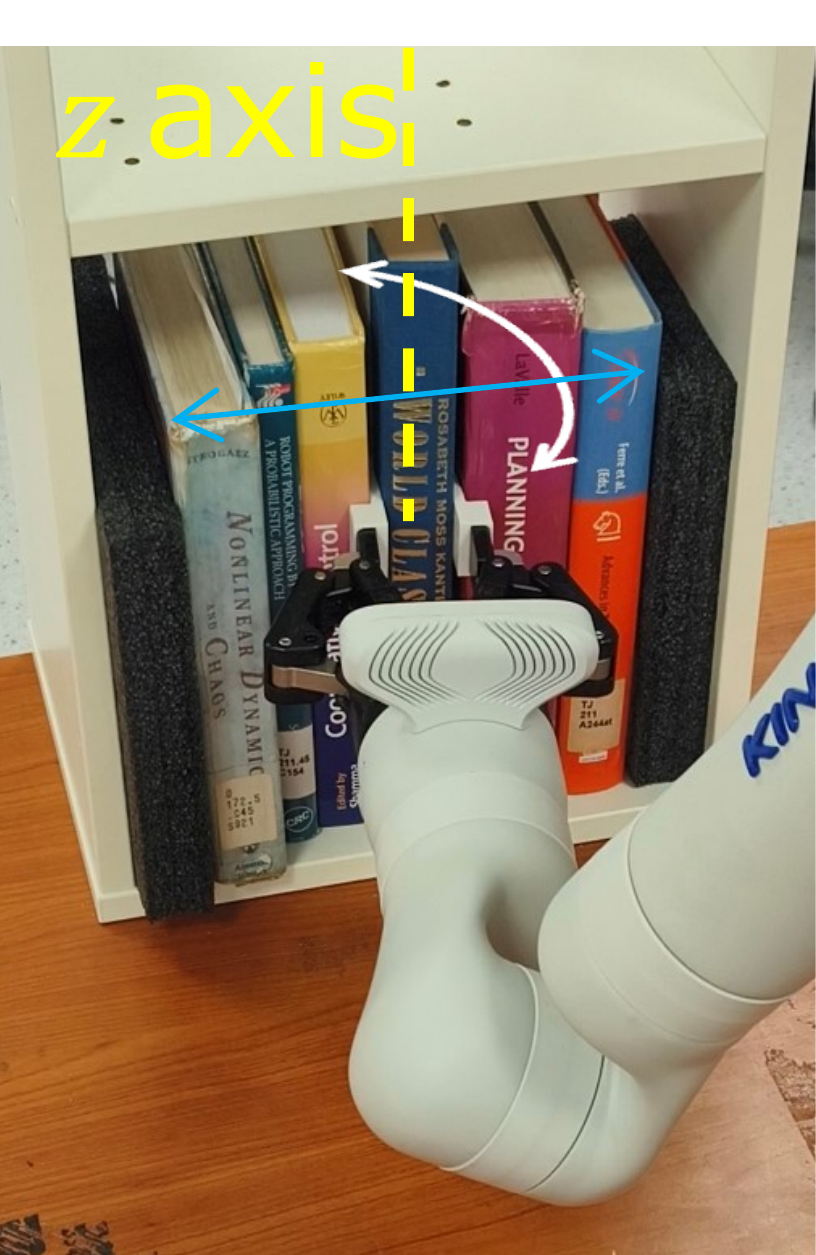}
    \caption[] 
    {\small Stiffness determination: Manually put the book onto the shelf, then control the robot with a constant force or torque.}
    \label{fig:exp cali}
  \end{subfigure}
  \caption{Experiment setup and stiffness determination.} 
\end{figure}

Fig. \ref{fig:exp setup} illustrates experimental setup. We use a Kinova to manipulate a book, with two fingers grasping the book. The Kinova is controlled in 2D. The books are manually placed on the bookshelf, where the remaining gap $w_2$ is narrower than the manipulated book’s width $w_1$. Two pieces of foam are used to adjust the stiffness of the setup. The target position is manually guided and recorded in the Kinova robot's frame.

\vspace{-3mm}


\vspace{-5mm}
\subsection{Experiment: Determination of stiffness}

To accurately simulate the interaction with the books on the shelf, we first conduct a pre-calibration to determine their stiffness. As illustrated in Fig. \ref{fig:exp cali}, we place the book in the slot and apply a constant force of $20 \; N$ or a torque of $5 \; Nm$ to the robot. The resulting displacement or rotation of the book along the z-axis is recorded. From this, we determine the stiffness to be $k_{ix} = 350 \; N/m, \; k_{i\theta} = 20 \; Nm/rad$.


\subsection{Simulation: Policy improvement during BBO}

\begin{figure}[H] 
  \begin{subfigure}[b]{0.5\linewidth}
    \centering
    \includegraphics[width=0.8\linewidth]{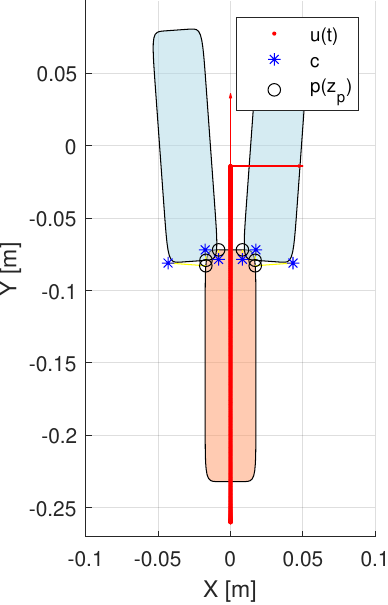}
    \caption{Initial condition, straight insertion results failure} 
    \label{fig7:a} 
    \vspace{4ex}
  \end{subfigure}
  \begin{subfigure}[b]{0.5\linewidth}
    \centering
    \includegraphics[width=0.8\linewidth]{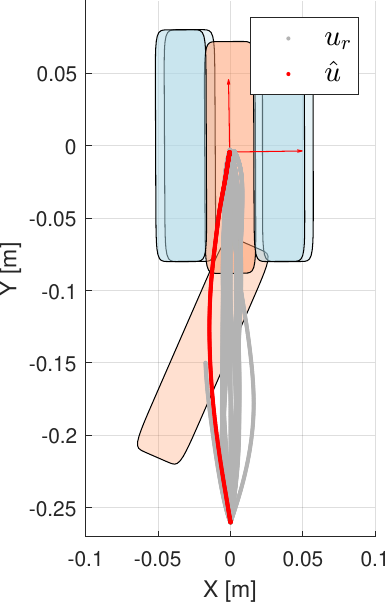}
    \caption{Iteration 1: $\phi=27.96$, push with front} 
    \label{fig7:b} 
    \vspace{4ex}
  \end{subfigure} 
  \begin{subfigure}[b]{0.5\linewidth}
    \centering
    \includegraphics[width=0.8\linewidth]{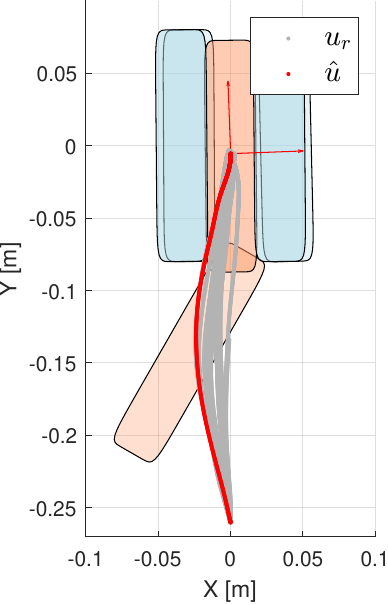}
    \caption{Iteration 3: $\phi=23.65$, push with side} 
    \label{fig7:c} 
  \end{subfigure}
  \begin{subfigure}[b]{0.5\linewidth}
    \centering
    \includegraphics[width=0.8\linewidth]{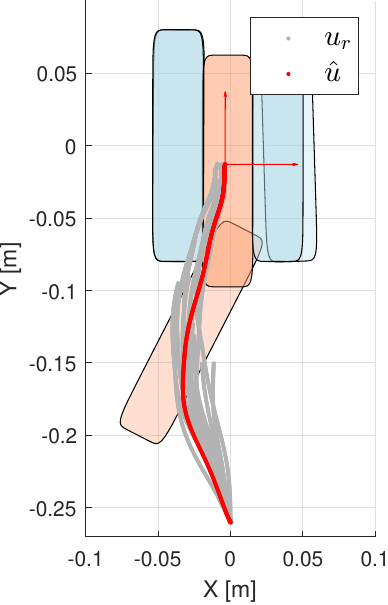}
    \caption{Iteration 10: $\phi=22.06$, push with side} 
    \label{fig7:d} 
  \end{subfigure} 
  \caption{All variations in iteration.}
  \label{fig7} 
\end{figure}

In this section, we demonstrate how the policy is improved based on our cost function $\phi$. We select a control stiffness $k_{cx} = k_{cy} = 800 \; N/m, \; k_{c\theta} = 20 \; Nm/rad$, ensuring they are higher than the external stiffness while remaining within the robot's capabilities. Each iteration consists of $R = 15$ rollouts.
The parameter of DMP is initialized as $\V \Theta = \V 0$, which corresponds to a straight-line trajectory, as shown in Fig.\ref{fig7:a}. A straight insertion policy fails due to the crowded shelf. Meanwhile, the corner of book $\V c$ and proxy on another book $\V p$ are plotted, with a yellow line connecting these proxy-corner pairs. 
We observe overlap between the manipulated book and the neighboring book (blue), where the corner of the manipulated book penetrates into the neighboring book. This penetration results in a high stiffness due to the nonlinear stiffness model (Eq. \ref{eq: stiffness}), causing the book stuck.

\begin{figure}[!h]
\centering
\includegraphics[width=0.7\linewidth]{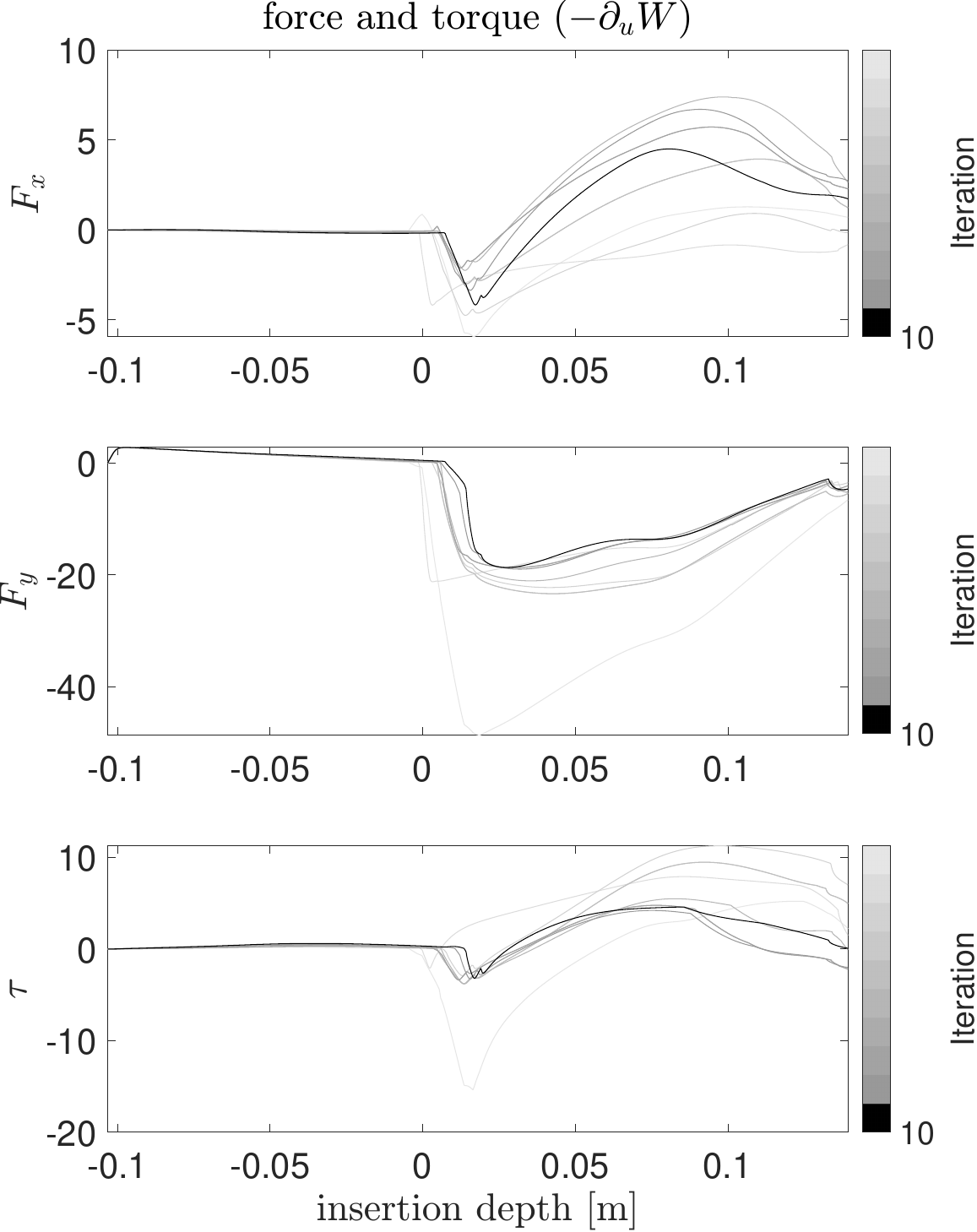}
\caption{Optimal control force in each iteration, iteration is shown as gray value.}
\label{fig:iteration_force}
\end{figure}

\vspace{-3mm}
The three subfigures in Fig. \ref{fig7} illustrate the policy improvement across iterations. The optimal policy, $\hat{\V u}(t)$, is selected based on the cost function $\phi$, while non-optimal policies are shown in grey. We display the entire trajectory, $\V u(t)$, along with two states of the books: the initial contact and the final state.
Remarkably, after just one iteration (15 explorations), the robot manages to insert the book with a tilt angle. However, this policy results in a large contact force. According to our methodology, the control force can be calculated as $-\partial_{\V u} W$. In Fig. \ref{fig:iteration_force}, we plot the three dimensions of this control force. The grey value of the trajectories corresponds to the iteration number, allowing us to observe policy improvement over iterations. Early iterations are represented with lighter color, while the final optimal policy is shown in black. Notably, the initial iterations exhibit substantial contact forces.

In Fig.\ref{fig7:c} and Fig.\ref{fig7:d}, we observe that at the initial contact, the book tilts more, using its long side (the front page) to push against the neighboring book on the left, rather than the short side as seen in iteration 1 in Fig. \ref{fig7:a}. 
This adjustment is reasonable, as contacting with the long side reduces the moment arm on the book, thereby decreasing the required control force and torque. This trend is evident in Fig. \ref{fig:iteration_force}, where the optimal policy (the black curve) is closest to zero. Additionally, we notice a rapid decrease in contact force over the first few iterations, with the policy converging in later iterations.
Note that some policies fail to reach the target, indicating that the ODE was terminated either due to a haptic obstacle or because $\V z$ moved out of the workspace.

\subsection{Experiment: Qualitative results analysis}

\begin{figure}[!h]
\centering
\includegraphics[width=\linewidth]{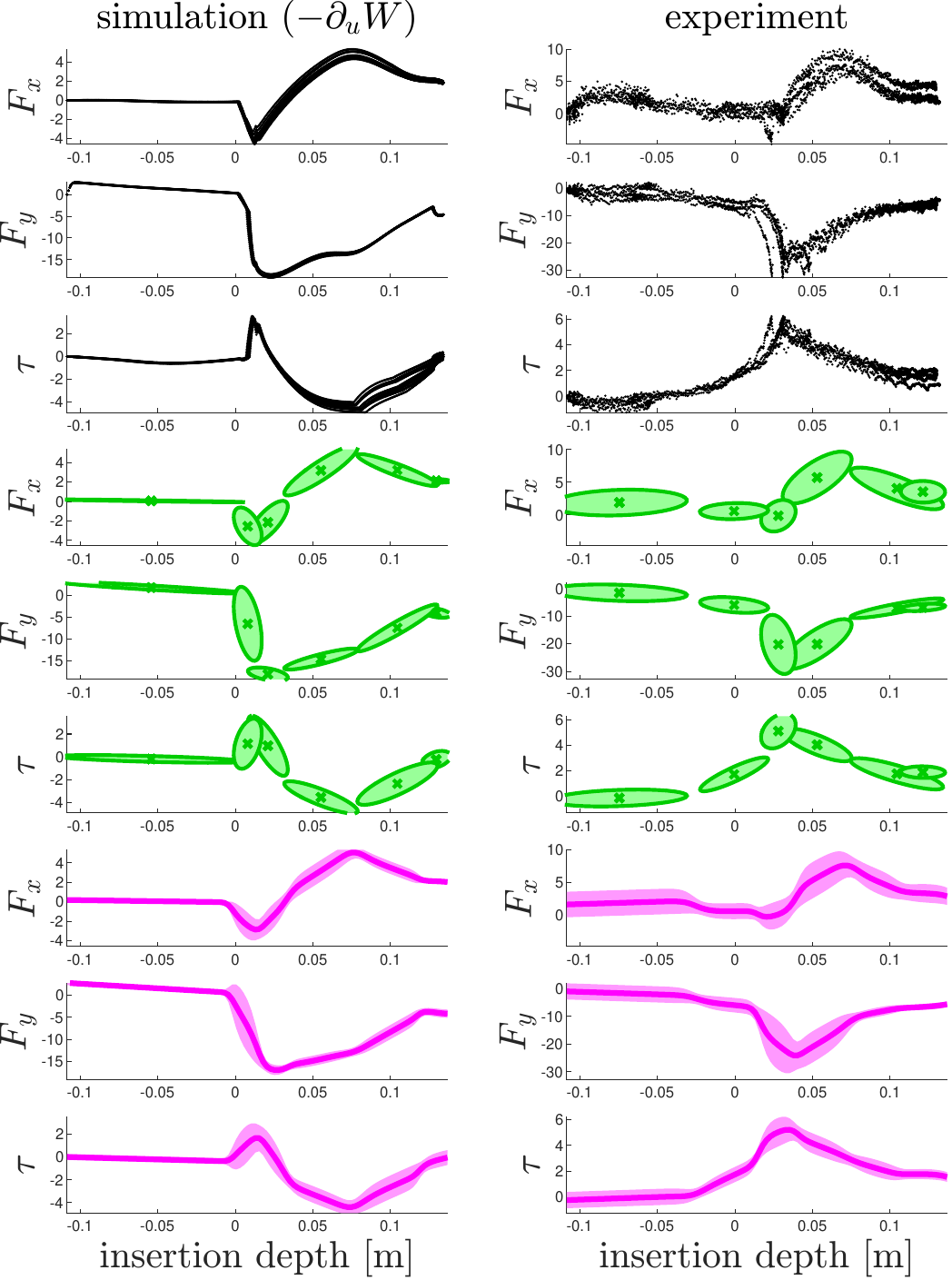}
\caption{Simulated contact force $-\partial_{\V u} W$ (with variation) v.s. experimental contact force, all in book frame. Black: original data. Green: cluster by GMM. Pink: regression by GMR. }
\label{fig:sim2real}
\end{figure}

To represent control noise, we introduce similar Gaussian variations in BBO to the optimal control policy parameters, $\hat{\V \Theta}$. This results in similar contact forces and torques, represented by the black lines in Fig. \ref{fig:sim2real}. Subsequently, we utilize a Gaussian Mixture Model (GMM), shown in green, and Gaussian Mixture Regression (GMR), in pink to statistically analyze the optimal policy. Due to the quasi-static assumption, we plot insertion depth \cite{whitney1982quasi} versus force instead of time.
On the right side, we repeat the optimal policy in the real world 5 times. Because of control noise and the need to manually reposition the books on the shelf, there are minor differences in each trial. The original force data is plotted in black, with the GMM and GMR analysis following the same color scheme.


Although the insertion policy is discovered by our framework, using statistical clustering methods GMM, we can \textit{interpret} its strategy through 5 distinct phases.
\begin{itemize}
    \item Before contact: Both force and torque feedback are zero as the robot adjusts the book’s pose. In simulation, this is represented by a single Gaussian (the first green ellipse), while in real-world data, two Gaussians are required for the cluster. This discrepancy may arise from the joint torque sensor capturing the robot's inertial effects.
    \item Push aside: The book tilts to push the neighboring book aside, requiring a torque and resistance force, hence all of them start increasing.
    \item Push forward: Once the gap becomes wide enough for insertion, the robot begins pushing forward. During this phase, resistance force in y axis (push in direction) decreases, but during alignment, force in x axis is varying.
    \item Slide in: The robot slides the book fully into the gap, with the force and torque decreases as the book’s orientation becomes more aligned. 
    \item Finish: All the force and torque converges in the end.
\end{itemize}
Thus, our framework autonomously discovers an explainable insertion strategy. Meanwhile, we notice the trend of contact force and torque are similar.

\subsection{Experiment: Variation of initial position}
Referring to Fig. \ref{fig:BBO and Generalization}, 
we uniformly vary the initial position of the book and apply the same BBO framework to evaluate the robustness of our algorithm. Each optimal control trajectory, $\V u(t)$, is then implemented in the real world and tested five times per case.
The results in Table \ref{table_result} demonstrate the effectiveness and robustness of our framework. In all cases, the framework successfully identified the wedging-in policy, even when starting from different positions. Furthermore, the integration of impedance control effectively compensated for minor discrepancies between the simulation and the real world, which were caused by the manual placement of the books.
\begin{table}[!h]
\renewcommand{\arraystretch}{0.9}
\caption{Variation of book's initial position.\label{table_result}}
\centering
\resizebox{\linewidth}{5mm}{
\begin{tabular}{cccccc}
\toprule
\textbf{Initial position in x axis (m)} & 0 & 0.025 & 0.050 & 0.075 & 0.100 \\
\midrule
\textbf{Successful Result} & 5/5 & 5/5 & 5/5 & 5/5 & 5/5 \\
\bottomrule
\end{tabular}}
\end{table}

\subsection{Experiment: Variation of external stiffness}
We vary the external stiffness $\V K_i$ on the neighboring books to test our framework. We first conduct a simulation where the external stiffness on the left and right sides is varied with ratios from 1:2 to 1:4. In all cases, the framework successfully determined the wedging-in policy, confirming the conclusion: the robot initially pushes the softer side upon first contact to initiate the wedging-in process.
\begin{figure}[H]
\centering
\includegraphics[width=0.9\linewidth]{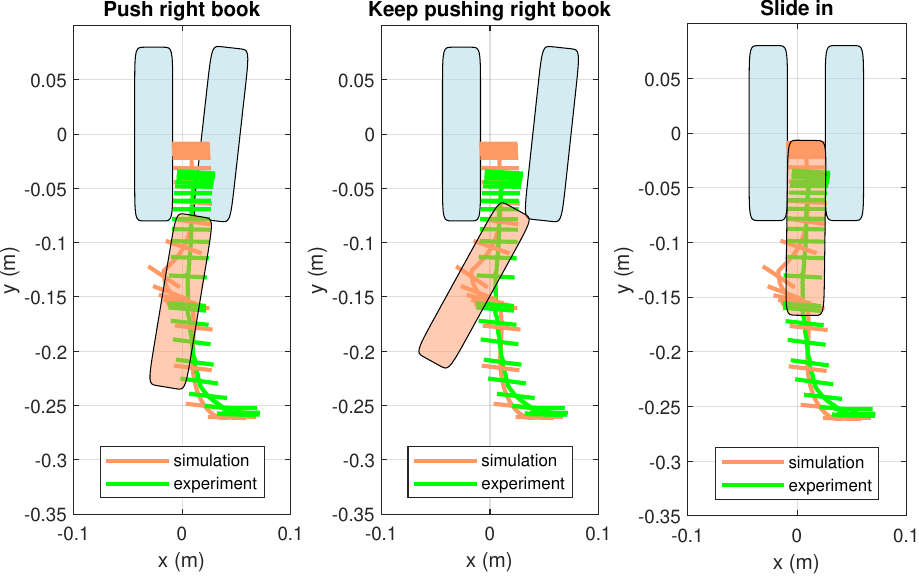}
\caption{Representative states when the left side is rigid. Our framework identifies the need to apply greater force on the right side (common books) until a sufficient gap is created.}
\label{fig:rigid}
\end{figure}

In real world, We set up 2 scenarios for this test: with the left side as the bookshelf edge and the right side as the edge.
Our framework still discovers a wedging-in policy, as shown in Fig.\ref{fig:rigid}. One key observation is that the manipulated book does not push against the left side (bookshelf edge), as the large stiffness results in a high value for $\V G(\V z^*(\V u), \V u)$, leading to a high cost. This policy can be \textit{interpret} by 3 phases.

In the left subfigure, the book pushes against the right-side book to adjust its orientation. In the middle subfigure, the robot continues pushing to create sufficient space. Finally, in the right subfigure, the book slides smoothly into the gap without pressing on the left side.
This strategy proves effective in the real world, even though the book temporarily sticks at the shelf front (seen by the accumulated dash line around $y=-0.15$). 
However, a difference emerges during the second phase: while our SQ model results in a tilt angle of the manipulated book due to its rigid, quasi-static assumptions, real-world interactions show deformation and slip at the spine of the right adjacent book as the robot keeps pushing.
Despite this discrepancy, our strategy successfully inserts the book into the shelf.
The results are summarized in Table \ref{table_result_stiff}. The outcomes demonstrate that our method successfully identifies effective strategies based on the specific edge conditions.


\begin{table}[H]
\centering
\caption{Variation of external stiffness.\label{table_result_stiff}}
\begin{tabular}{@{}p{3cm}p{2cm}p{2cm}@{}}
\toprule
\textbf{Edge of bookshelf} & Left Edge & Right Edge  \\
\midrule
\textbf{Successful Result} & 5/5 & 5/5  \\
\midrule
\textbf{Discovered strategy} & Push right book & Push left book \\
\bottomrule
\end{tabular}
\end{table}

\subsection{\HL{Analysis: comparative study}}
\HL{
We now conduct a comparative study to analyze the components in our framework.

\subsubsection{Manipulation potential}
Our framework only requires defining a manipulation potential that captures all contacts between objects and the impedance control. With this formulation, both system states and contact points are implicitly defined and can be solved through an ODE (Eq. \ref{eq:adaptive ODE}). In contrast, many existing approaches in manipulation rely on explicitly defining contact points and specifying contact constraints at each location \cite{pang2023planning,doshi2022manipulation}.

\subsubsection{Proxy capturing contact}

Accurate modeling of contact requires computing the closest distance and contact points between objects. Direct computation methods (e.g., Euler-Lagrange formulations or swarm particle methods) can be computationally expensive and complex. To address this, we introduce a virtual “proxy” point, which transforms the problem of determining contact point into a continuous ODE, significantly simplifying and accelerating computation. The proxy is a zero-mass point that slides along the surface of the object, representing the contact point. This idea also leverages high-dimensional space advantages, similar to kernel methods \cite{hofmann2008kernel}, and is supported by findings in \cite{udwadia2021use}.

Moreover, this ODE-based proxy model enhances stability. In certain cases, computing the closest point directly may result in discontinuities \cite{le2024contact}. For instance, when an object penetrates deeper and the true closest point $\V p_c$ shifts abruptly, causing sudden changes in contact force and instability, as shown in the Fig. \ref{fig:SQ_proxy_after_ct}. Our proxy with nonlinear stiffness avoids such discontinuities, offering a smooth, stable, and efficient contact representation.

\subsubsection{Adaptive ODE}
The variable $\V z$ in our model is implicitly defined. In principle, one could use numerical solvers such as MATLAB’s fsolve to compute the equilibrium $\V z^*$. However, as shown in \cite{campolo2025geometric}, our system exhibits multi-branched manifolds, meaning that for the same control input $\V u$, multiple valid equilibrium states $\V z$ may exist.
We address this multi-branch issue by introducing a haptic obstacle (Eq. \ref{eq: Haptic Obstacle}) to constrain the solution to same branch. A clear example of this phenomenon is given in the inverted pendulum task shown in Fig. 4 of \cite{campolo2025geometric}, where same control $\V u$ can yield different system states $\V z^*$ based on initial guess of “fsolve”. There exist two branches of the manifold. At the same value of $\V u$, multiple solutions for $\V z^*$ exist, corresponding to different branches of the manifold. Without an adaptive ODE solver that ensures the solution remains on the same branch, the solution may jump between branches, leading to incorrect behavior.

In the context of the book insertion task, without a wedging-in strategy, even the control $\V u$ is inside bookshelf, the book will be still stuck in front of bookshelf. Hence, directly using fsolve may lead to different solutions. Our adaptive ODE ensures convergence to same branch while maintaining stability.
}


\section{Conclusion}
We present a haptic manipulation planning framework in a naturally separated configuration space, using an augmented algorithm to explore the implicit manifold, compute contact points, and measure haptic distance via a haptic metric. Verified in crowded book insertion, our framework uses BBO to discover strategic insertion policies under varying stiffness and initial conditions, reducing contact force and torque during optimization. Our SQ and proxy models effectively capture contact interactions, closely aligning with real scenarios except for deformation and slip effects.
We use the spine side of the book for insertion, as inserting with the page side could lead to failure due to the cover. 
Additionally, we simplify book insertion as a planar task, which assumes bookshelf support and limits its applicability to certain scenarios.
\HL{As a future direction, we plan to integrate visual perception into our framework, enabling automatic estimation of object geometry using SQ fitting.}
Additionally, we will demonstrate the numerical optimality of our policies in the real world.


\bibliographystyle{ieeetr}
\bibliography{ref}

\end{document}